\documentclass[runningheads]{llncs}
\usepackage[T1]{fontenc}
\usepackage{graphicx,verbatim}
\usepackage{algorithm2e}
\usepackage{amsmath}
\usepackage{amssymb}
\usepackage{booktabs}
\usepackage{multirow}
\usepackage{array}
\usepackage{subfig}
\usepackage{hyperref}
\hypersetup{
    colorlinks=true,
    linkcolor=green,
    urlcolor=magenta
}
\usepackage{diagbox}
\usepackage{float} 
\begin{document}
\title{Policy to Assist Iteratively Local Segmentation: Optimising Modality and Location Selection for Prostate Cancer Localisation}
\titlerunning{Policy to Assist Local Segmentation}
%
%
\author{
Xiangcen Wu\inst{1,2}\and
Shaheer U. Saeed\inst{1,2} \and
Yipei Wang\inst{1,2} \and
Ester Bonmati Coll\inst{1,2,3} \and
Yipeng Hu\inst{1,2}
}
\authorrunning{Xiangcen Wu et al.}
%
\institute{Department of Medical Physics and Biomedical Engineering, University College London, London, UK \and
UCL Hawkes Institute, University College London, London, UK \\
\and
School of Computer Science and Engineering, University of Westminster, London, UK\\
\email{xiangcen.wu.21@ucl.ac.uk}}
\maketitle              
\begin{abstract}
Radiologists often mix medical image reading strategies, including inspection of individual modalities and local image regions, using information at different locations from different images independently as well as concurrently. In this paper, we propose a recommend system to assist machine learning-based segmentation models, by suggesting appropriate image portions along with the best modality, such that prostate cancer segmentation performance can be maximised. Our approach trains a policy network that assists tumor localisation, by recommending both the optimal imaging modality and the specific sections of interest for review. During training, a pre-trained segmentation network mimics radiologist inspection on individual or variable combinations of these imaging modalities and their sections - selected by the policy network. Taking the locally segmented regions as an input for the next step, this dynamic decision making process iterates until all cancers are best localised. We validate our method using a data set of 1325 labelled multiparametric MRI images from prostate cancer patients, demonstrating its potential to improve annotation efficiency and segmentation accuracy, especially when challenging pathology is present. 
Experimental results show that our approach can surpass standard segmentation networks. Perhaps more interestingly, our trained agent independently developed its own optimal strategy, which may or may not be consistent with current radiologist guidelines such as PI-RADS. This observation also suggests a promising interactive application, in which the proposed policy networks assist human radiologists. The code is open-sourced and available at \href{https://github.com/XiangcenWu/ModalitySelection}{here}.

\keywords{Prostate Cancer  \and Reinforcement Learning \and Image Segmentation.}

\end{abstract}

\section{Introduction}

Accurate annotation of tumors in prostate cancer plays a crucial role in targeted biopsy \cite{siddiqui2015comparison,ahmed2017diagnostic,kasivisvanathan2018mri,RADTKE201587}, focal therapy \cite{ahmed2011focal,VALERIO2014732} and emerging MRI-targeted radiotherapy \cite{lagendijk2014mr}. High-quality annotations help in treatment planning, monitoring disease progression, and training deep learning models for automated detection \cite{duran2022prostattention,pellicer2022deep}. However, this process is time-consuming and labor-intensive for radiologists. They must analyse multi-parametric MRI scans in multiple aspects, for example, to accurately identify the tumor location and its boundary as well as to select the appropriate imaging modality. Since certain lesions or lesion areas may only be visible in a specific modality, switching between modalities is often necessary for cancer annotation. This is in contrast to existing deep learning models for prostate cancer detection \cite{pellicer2022deep}, which often adopt feed-forward neural networks with one-pass inference unable to model such a dynamic decision making process. Arguably, data-driven approaches may be capable of learning such a complex decision-making. In this application of localising prostate cancer, however, the performance has been highly limited due to practical constraints such as quality and quantity of the available labeled data \cite{jiang2023prostate}.

In this work, we focus on using reinforcement learning (RL) to detect cancer on MRI in contrast to conventional supervised cancer segmentation models. Firstly, it models the above-discussed dynamic decision making which optimises the agent's cancer detection strategy based on changes of state (e.g., here, the images, selected sections, MRI sequences and current segmentation). Secondly, it leverages the optimised trajectories of ``actions'' for improving interoperability. Thirdly, the agent we train can potentially surpass the supervised segmentation model due to exploration and trial-and-error based learning as opposed to only learning to imitate human labels. Additionally, RL enables multi-step decision processes, such as selecting the most informative imaging modality and annotating cancerous regions portion by portion. In practice, the agent can be integrated to provide actionable guidance to radiologists while labeling cancerous region on MRI, rather than used alone to provide a final prediction. Although not discussed further, potential wider applications include providing feedback in clinician training, interactive assistance and enhancing transparency, trustworthiness and repeatability in clinical practice.

Outside of prostate cancer detection, previous work also explored RL approaches for medical image computing applications, such as anatomical segmentation \cite{stember2020deep} and pathology detection \cite{gayo2022strategising}. However, most of these existing RL-based algorithms aimed at improving detection performance, without explicitly modeling a human actionable environment. 
In contrast, we propose a policy network for RL agent, and a second segmentation network, separately developed and trained as a `simulated radiologist'. The simulated radiologist is trained to recognise and process different input modalities at various portions of the image volume. These image portions could be represented by image patches or a subset of slices in a volumetric image. These segmentation outputs are then used to give reward to the policy network of our agent and optimises its strategy. We highlight that optimisation of the policy network relies on rewards derived from segmentation accuracy across different modalities and portions, as provided by the simulated radiologist. This process of trial and error allows our agent to explore different paths and therefore potentially discovers different strategies that may result in different and better cancer localisation performance.

To summarise, we introduced RL for cancer detection in MRI, contrasting with conventional supervised segmentation approaches, where the iterative learning process enables the agent to discover interesting workflows for modality selection and image portions to be focused on. We proposed a policy network for RL that interacts with a separately trained segmentation network that serves as a surrogate reward function, improving detection accuracy across different MRI modalities. Clinical data is used for a real-world application, highlighting its potential integration into radiology workflows. This iterative learning process enables the agent to discover interesting paths for cancer detection. Open source code is provided for reproducibility and further research \href{https://github.com/XiangcenWu/ModalitySelection}{here}.


\section{Method}

We formulate the prostate cancer localisation problem as a Markov Decision Process (MDP) where a radiologist first selects the imaging modality, and portion of the anatomy, which are likely to yield the best local segmentation. The radiologist then segments the cancer on the selected portion of the anatomy using the selected modality. This process is repeated until all the cancerous regions have been segmented for all portions of the anatomy. We formulate this decision process as a RL problem which aims to find the best patient-specific strategy for imaging modality selection such that cancer segmentation performance is maximised. To this end, we simulate the radiologist's workflow for segmenting cancer on a given medical image, and discover an optimal strategy using RL.

\subsection{Simulated radiologist during training}
\label{Simulated radiologist during training}
A multi-modal volumetric image sample $x \in \mathbb{R}^{H \times W \times D \times C}$ is composed of $C$ co-registered volumes of size $H \times W \times D$, each corresponding to a different imaging modality. An image potion\footnote{Without loss of generality, an image portion in this study refers to a subset of slices in a cropped single-modality image, cropped to contain at least the prostate gland. Further experimental details are described in Sec~\ref{Comparison and ablation studies}} is denoted as $x_p \in \mathcal{X}_p \in \mathbb{R}^{h \times w \times d \times 1}$, where $h \times w \times d$ denote the dimensions of the image portion, with $h\leq H$, $w\leq W$ and $d\leq D$, $1$ denotes the single modality that is being used for segmentation, and $\mathcal{X}_p$ denotes the distribution of image portions. To simulate radiologist segmentation during RL training, we construct a deep neural network $f(\cdot; \theta): \mathcal{X}_p \rightarrow \mathcal{Y}_p$, which predicts a segmentation $y_p \in \mathcal{Y}_p$ for an image portion $x_p \in \mathcal{X}_p$, and has weights $\theta$, where $\mathcal{Y}_p$ denotes the space of possible segmentation labels.

In practice, the segmentation network $f(\cdot; \theta): \mathcal{X} \rightarrow \mathcal{Y}$ is trained on full-sized multi-modality image volumes $x \in \mathbb{R}^{H \times W \times D \times C}$ rather than on cropped image portions. Each training sample contains $C$ modalities and is configured to simulate radiologist viewing conditions: single modality scenario, or view all modalities. To simulate these settings, a masked input $\tilde{x}$ is generated by zeroing out the channel corresponding to the modality that is not viewed (for single-modality configurations), or by retaining all channels when simulate viewing all modalities. The predicted full-volume segmentation $y = f(\tilde{x}; \theta^*)$ is later divided into spatial sub-volumes $y_p \in \mathcal{Y}_p$, consistent with cropped inputs $x_p \in \mathcal{X}_p$. This segmentation network $f(\cdot; \theta)$ is trained using a loss function $\mathcal{L}(\cdot): \mathcal{Y} \times \mathcal{Y} \rightarrow \mathbb{R}_{\geq0}$, which measures similarity of the neural network segmentation $f(\tilde{x}; \theta)$ compared to a human label $\hat{y}\in\mathcal{Y}$. For notational convenience, we use image portions segmentation $f(\cdot; \theta): \mathcal{X}_p \rightarrow \mathcal{Y}_p$ in the following section. The optimisation of segmentation network parameter $\theta$ is given in:
$$
\theta^* = \arg \min_{\theta} \mathbb{E}_{\tilde{x}, \hat{y} \sim \mathcal{X}, \mathcal{Y}} \left[ \mathcal{L}(f(\tilde{x}; \theta), \hat{y}) \right].
$$

\subsection{Markov Decision Process for navigating cancer localisation}

We formulate the radiologist's segmentation workflow, i.e., the image portion and modality selection followed by segmentation, as an MDP such that we can use a RL policy to learn the optimal selection strategy and workflow. In this formulation the radiologist observes the environment and then takes an action, which influences the environment leading to a reward signal, which is used to optimise the workflow.

\textbf{Environment:} In the segmentation workflow, the environment consists of the simulated radiologist $f(\cdot; \theta^*)$ and the image to be segmented $x$ from which image portions for single modalities $x_p\in \mathcal{X}_p$ can be derived. Since we are modeling a decision process, each step in the decision is denoted by a time step $t$, which will be used as a subscript for variables in further analyses.

\textbf{States:} The observed state $s_t$ consists of image $x$, as well as the corresponding segmentation $y_t$ collected by previous time steps. The state is thus denoted as $\{ x, y_{t}\} = s_t\in\mathcal{S}$, where $\mathcal{S}$ is the state space.

\textbf{Actions:} The action consists of a portion selection action $a_{p,t} \in \{1, ..., P\}$, selecting one of $P$ portions, and a modality selection action $a_{m,t} \in \{1, ..., C\}$, selecting one of $C$ modalities. So, the action is denoted as $\{a_{p,t}, a_{m,t}\}=a_t \in \mathcal{A}$, where $\mathcal{A}$ is the action space. In practice, this action space is represented as a discrete space of dimension $P \times C$, which allows portion and modality selection with a single discrete value.

\textbf{Transitions:} The transition to the next state (illustrated in Fig~\ref{The overall algorithm}) $s_{t+1}=\{x, y_{t+1}\}$ happens as the action $a_t$ leads to the selection of a new portion and modality with $a_{p,t}$, giving $x_{p,t+1}$. This can then be used to compute $y_{p,t+1}=f(x_{p,t+1}; \theta^*)$. The updated prediction \( y_{t+1} \) is obtained by replacing the region \( p \) in the previous prediction \( y_t \) with the newly computed \( y_{p,t+1} \), while keeping the rest of \( y_t \) unchanged. In other words, \( y_{t+1} \) is a partial update of \( y_t \), where only the selected portion \( p \) is revised based on the latest inference. The state transition distribution ${\rho}: \mathcal{S} \times \mathcal{S} \times \mathcal{A} \rightarrow [0, 1]$ denotes the probability of the next state $s_{t+1}\in \mathcal{S}$ given the current state $s_t\in\mathcal{S}$ and action $a_t\in\mathcal{A}$.

\textbf{Rewards:} The rewards are constructed to quantify cancer segmentation performance through a reward function $r(\cdot): \mathcal{S} \times \mathcal{A} \rightarrow \mathbb{R}$ where $R_t = r(s_t, a_t, s_{t+1})$ denotes the reward at time-step $t$. With $s_t=\{x, y_{t}\}$ we compute segmentation loss to the ground-truth label as $\mathcal{L}(y_{t}, \hat{y}_{t})$. The action $a_t$ transitions the state to $s_{t+1}=\{x, y_{t+1}\}$ as described in the ``Transitions'' paragraph above. This is then used to compute the reward $R_{t} = \mathcal{L}(y_{t}, \hat{y}) - \mathcal{L}(y_{t+1}, \hat{y})$.

\begin{figure}[htbp] 
  \centering 
  \includegraphics[width=1\textwidth]{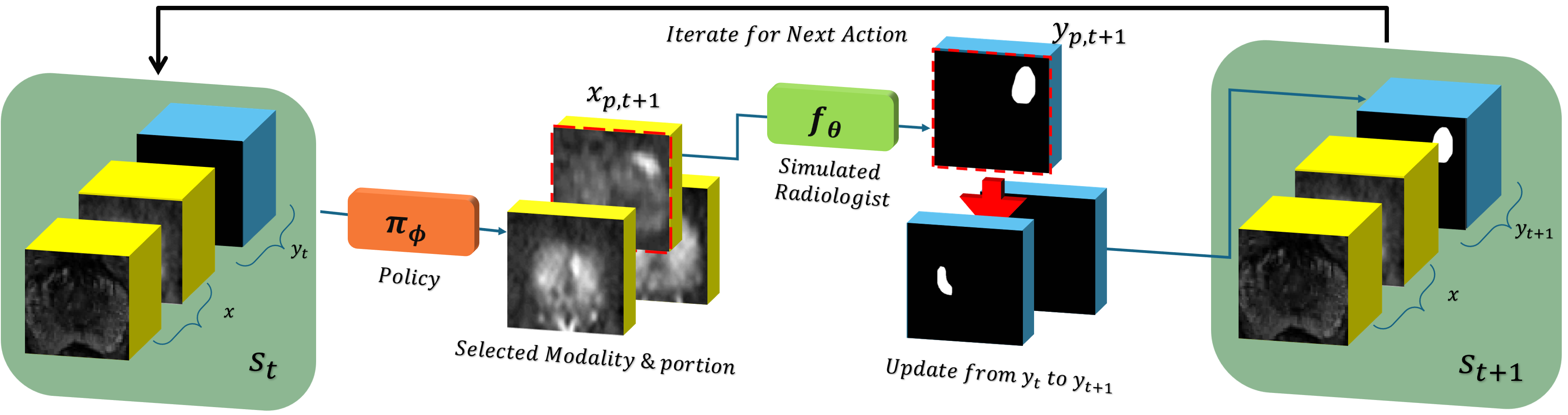} 
  \caption{\textbf{MDP State Transition:} At timestep $t$, the state $S_t = \{x, y_t\}$ includes the multi-modal image $x$ and current segmentation $y_t$. The policy $\pi_\phi$ selects a modality and region ($x_{p,t+1}$), which is segmented by $f_\theta$ to produce $y_{p,t+1}$. The updated mask $y_{t+1}$ defines the next state $S_{t+1}= \{x, y_{t+1}\}$. $S_{t+1}$ is then fed back into the policy $\pi_\phi$ continuing the iterative decision-making process.} 
  \label{fig:MDP} 
\end{figure}
\subsection{Optimising radiologist segmentation  strategies using RL}
The policy $\pi_{\phi}(a\mid s):\mathcal{S}\times\mathcal{A}\rightarrow[0,1]$ represents the probability of performing action $a_t$ for portion and modality selection, given $s_t$. We can accumulate rewards starting from time-step $t$, following the policy, using $Q^{\pi_{\phi}}(s_t, a_t) = \sum_{k=0}^{T}\gamma^kR_{t+k}$, where $\gamma\in[0,1]$ is a discount for future rewards. With the MDP interactions of the environment with the policy we observe a trajectory $(s_1, a_1, R_1, s_2, a_2, R_2, \ldots,  \\s_T, a_T, R_T)$. The objective in reinforcement learning is to optimise the policy parameters $\phi$ to maximise expected cumulative reward. Optimising over a training set of patient cases enables generalisation to new patients, even when ground-truth labels are unavailable during testing. If we consider a parameterised policy $\pi_\phi$ that generates this trajectory, then the optimisation objective for RL is to learn the optimal policy parameters:

$$\phi^* = \arg \max_{\phi} \mathbb{E}_{\pi_\phi}[Q^{\pi_{\phi}}(s_t, a_t)]$$

Optimising this across patient cases allows generalisability to new patients at test-time even when a ground-truth label is not available for reward computation.

\RestyleAlgo{algoruled} 
\begin{algorithm}[hbt!]
\caption{Radiologist segmentation workflow optimisation}\label{The overall algorithm}
\SetAlgoLined 
Randomly initialise policy network $\pi_{\phi}(a,s)$ with parameter $\phi$\;
Optimise $\theta$ for simulated radiologist $f(\cdot; \theta^*)$ \;
\While{ not converged, \do}{
\While {not done}{
    Randomly sample patient case from $\mathcal{D}^{rl}$\;
    Get $s_0=\{ x, y_{0}\}$ where $y_{0}$ is a full zero tensor\;
    
    \For {$episode=1, \boldsymbol{T}$}{
        
        Sample action $a_t\sim \pi_{\phi}(\cdot, s_t)$ according to the current policy\;
        Execute action $a_t$\;
        Observe new state $s_{t+1}$\;
        Observe reward $R_t$ based on reward function $r(s_t, a_t, s_{t+1})$\;
        Collect sequence $(s_t, a_t, R_t)$\;
    }
    Calculate discounted cumulative return $u_t$ for all $T$ within full episode\;
    Update parameter: $\phi \gets  \phi - \frac{1}{T} \sum_{t=1}^{T} u_t \cdot \nabla_{\phi} \ln \pi(a_t|s_t; \phi)$\;
}
}
\end{algorithm}

\section{Experiment}

The dataset is a large dataset collected as part of multiple ethically approved clinical studies conducted by institutions in Miami, FL \cite{miami_data}. The combined dataset $\mathcal{D}$ is partitioned into three subsets: one for training a segmentation model ($\mathcal{D}^{seg}$), another for training a reinforcement learning policy ($\mathcal{D}^{rl}$), and a holdout set for performance evaluation ($\mathcal{D}^{h}$). $\mathcal{D}^{seg}, \mathcal{D}^{rl}, \mathcal{D}^{h}$ has 925, 300, 100 examples respectively. These subsets satisfy $\mathcal{D} = \mathcal{D}^{seg} \cup \mathcal{D}^{rl} \cup \mathcal{D}^{h}$. All image and modality volumes are center-cropped based on the radiologist annotation of the prostate gland, intensity-normalised, and resampled to an image size of $128 \times 128 \times 32$. In our implementation, the agent is allowed to select between three options: T2-weighted (T2) and high b-value diffusion-weighted (DW) MRI, or viewing both modalities together.

\subsection{Comparison and ablation studies}
\label{Comparison and ablation studies}
SwinUNETR-v2 \cite{swinunetr-v2} is used as our segmentation network to simulate radiologist. The segmentation network takes channel-wise concatenation of $T2w$ and $DW$ image volume: $x\in\mathbb{R}^{h\times w\times d\times 2}$. If the agent selects to view both modalities, both channels are provided; otherwise, only the selected modality is retained, while the other channel is zeroed out as described in Sec~\ref{Simulated radiologist during training}. During training, each input is randomly configured with equal probability to one of three settings: $T2w$ channel masked, $DW$ channel masked, or both channels unmasked. To enhance robustness, we apply data augmentation techniques, including random affine transformations, Gaussian smoothing, Gaussian noise, and contrast adjustment, each with a probability of 0.25. The segmentation network is trained for a fixed 50 epochs using the Adam optimiser with a learning rate of $1e^{-4}$ and optimised with the Dice loss function.

We test two reinforcement learning algorithms: REINFORCE \cite{reinforce} (Algo. \ref{The overall algorithm}), Group Relative Policy Optimization (GRPO)~\cite{guo2025deepseek}. GRPO differs from REINFORCE in its use of a clipped surrogate objective inspired by Proximal Policy Optimization (PPO)~\cite{ppo}, along with a KL divergence regularisation term that encourages the learned policy to remain close to a reference policy $\pi_{\phi_{\text{ref}}}$. For the reference model, we use the REINFORCE ($\gamma = 0.5$), showed on Table~\ref{final results} and investigate the effect of varying the regularisation coefficient $\beta \in [0.1, 0.5, 1.0]$, which scales the KL divergence term $\beta*\mathbb{D}_{KL}(\pi_\theta \| \pi_{\text{ref}})$ as detailed in ~\cite{guo2025deepseek}. Different use of discounted factor $\gamma \in [0.3, 0.5, 0.8]$ for REINFORCE are also tested. For each image study, training samples are collected on-policy during learning, with 60 steps per image study. 300 imaging examples in $\mathcal{D}^{rl}$ are trained with 30 epochs. We divide the image volume into eight portions along the depth direction (4 slices per portion).


\begin{figure}[t] 
  \centering 
  \includegraphics[width=1\textwidth]{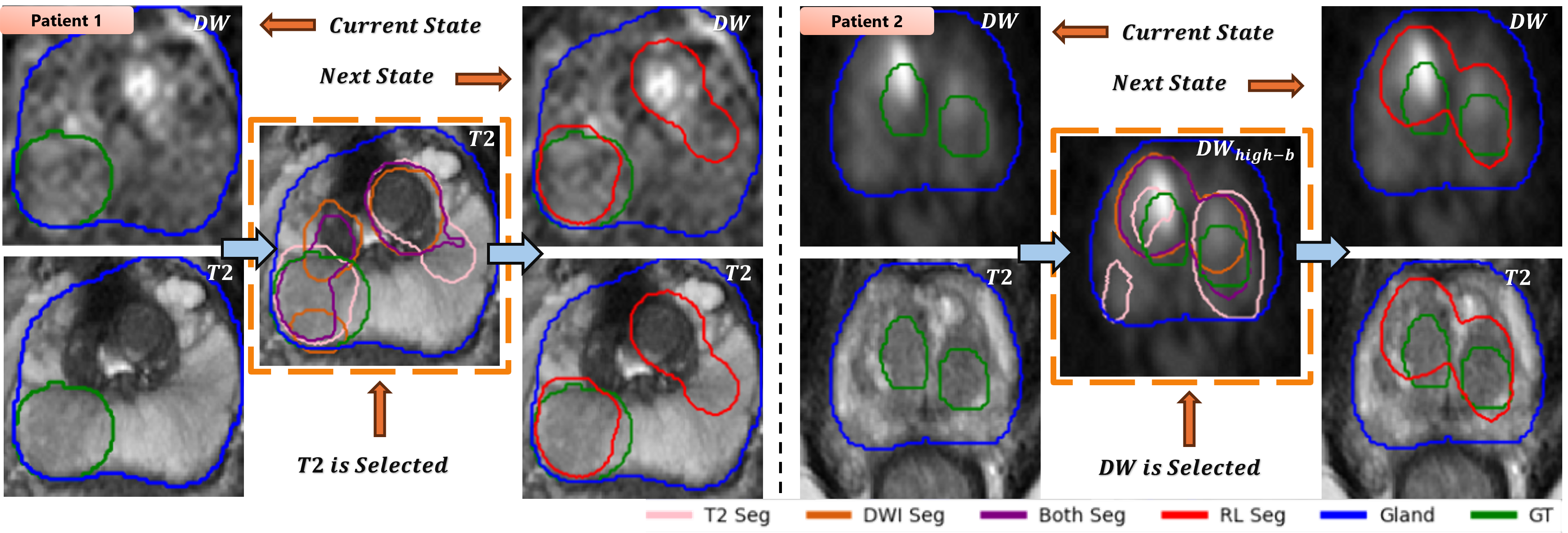}
  \caption{Illustration of two patient cases. The first column shows the current segmentation state; the second displays the selected modality and simulated radiologist annotation reading different modalities; the third shows the updated state after the agent's action.}
  
  \label{fig:Patient Examples} 
\end{figure}
\section{Results}
Presented Table~\ref{final results} shows the assessment of our agent's policy, we compare our agent's performance with the segmentation network. Unlike our RL-based approach, which follows a radiologist-like workflow by considering image portions and single modalities step by step, the segmentation network predicts the entire segmentation in a single pass without iterative decision-making. The agent's performance is evaluated by running the agent 10 time steps with it's action sampled by the policy network. We consider the Dice accuracy of current $y_{p,t}$ as the agent's performance for this patient. We also add segmentation model's performance on different modalities for comparison. $y_{p,t}$'s dice score has potential to surpass segmentation network on any of the three modalities as $y_{p,t}$ is the combination of different segmentation prediction by the segmentation network. 


\begin{table}[!ht]
\centering
\renewcommand{\arraystretch}{1.1}
\setlength{\tabcolsep}{8pt}
\begin{tabular}{llcccl}
\toprule
\multicolumn{2}{c}{\textbf{SwinUNETR}} & 
\multicolumn{2}{c}{\textbf{REINFORCE}} & 
\multicolumn{2}{c}{\textbf{GRPO}} \\
\cmidrule(r){1-2} \cmidrule(r){3-4} \cmidrule(r){5-6}
Modality & Dice  & $\gamma$ & Dice / Steps & $\beta$ & Dice / Steps \\
\midrule
$T2w$   & $0.29 \pm 0.17$ & 0.3 & $0.42 \pm 0.17$ / 9.40 & 0.1 & $0.43 \pm 0.17$ / 9.60 \\
$DW$   & $0.38 \pm 0.19$ & 0.5 & $0.43 \pm 0.17$ / \textbf{8.20} & 0.5 & $\textbf{0.44} \pm \textbf{0.17}$ / 9.50 \\
$Both$ & $0.40 \pm 0.18$ & 0.8 & $0.40 \pm 0.19$ / 10.10 & 1.0 & $0.43 \pm 0.16$ / 8.80 \\
\bottomrule
\end{tabular}
\caption{Comparison of segmentation performance (Dice score) and average steps across baseline segmentation models, REINFORCE variants, and GRPO variants.}
\label{final results}
\end{table}

To verify the impact of each component, we found that both the discount factor $\gamma$ in REINFORCE and $\beta$ in GRPO influences the training of our policy. Among the tested values, $\gamma=0.5$ results the optimal balance between steps taken and segmentation accuracy during inference , while $\beta=0.5$ in GRPO achieves the highest segmentation accuracy.

We also observed interesting agent behaviors in specific patient cases. For instance, as shown in Fig.~\ref{fig:Patient Examples}, patient 1 presents with a tumor located in the peripheral zone (PZ), yet the agent selected T2 for assessment. This choice contrasts with the PI-RADS v2.1 guidelines \cite{turkbey2019prostate}, which recommend high b-value diffusion-weighted imaging ($DW$) as the primary modality for PZ evaluation. A similar deviation is observed in Patient 2, where the agent choose to view a transition zone (TZ) lesion using DW, despite T2 being the preferred modality for TZ according to PI-RADS v2.1.

\begin{figure}[H] 
  \centering 
  \includegraphics[width=1\textwidth]{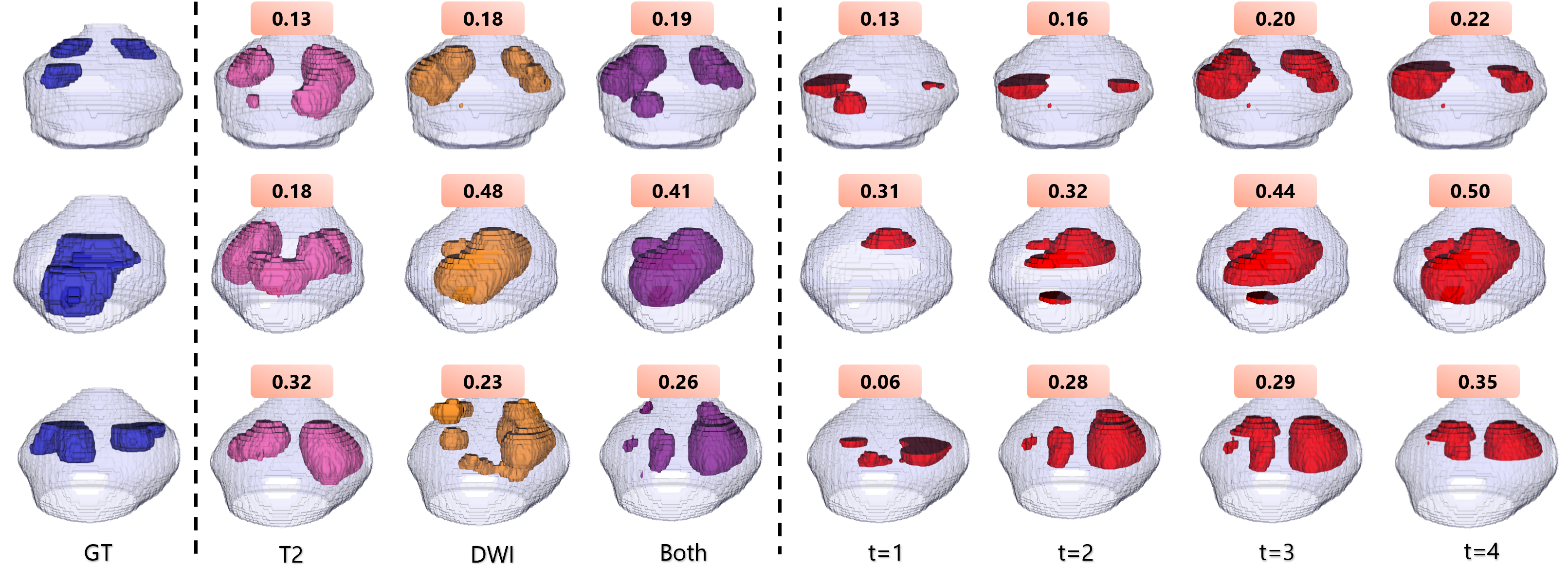} 
  \caption{Visualisation of ground truth (GT) and segmentation results using different input modalities (T2, DW, and Both), along with the reinforcement learning (RL) policy's detection over four time steps ($t=1$ to $t=4$). Dice coefficients, calculated with respect to the ground truth (GT), are displayed above.}
  \label{fig:miccai_3d} 
\end{figure}
\section{Conclusion}
In this work, we introduced a reinforcement learning (RL)-based framework to explicitly learn how radiologists inspect MRI images for prostate cancer localisation, by iteratively selecting and reviewing optimal imaging modalities and local anatomical regions. We introduce a policy network for the RL agent with a separately trained segmentation model acting as a simulated radiologist. Our approach allows agent for complex dynamic decision-making which differs from conventional supervised segmentation models. The experimental results demonstrate that the RL agent can improve segmentation accuracy and efficiency, adapting its decision-making strategy based on patient-specific imaging characteristics. Future research will investigate the feasibility to use the proposed policy networks working with radiologists, to test the capability in assisting a human-machine interactive workflow, otherwise unavailable in existing supervised localisation algorithms. Furthermore, the agent exhibits both human-like and novel behaviors, suggesting its potential to use in conjuction with radiologist to ease annotation workflows.
\section*{Acknowledgements}
This work is supported by the International Alliance for Cancer Early Detection, an alliance between Cancer Research UK [EDDAPA-2024/100014] \& [C73666/A31378], Canary Center at Stanford University, the University of Cambridge, OHSU Knight Cancer Institute, University College London and the University of Manchester; and National Institute for Health Research University College London Hospitals Biomedical Research Centre.

\bibliographystyle{splncs04}
\bibliography{references.bib}

\newpage

\end{document}